\documentclass[a4paper, 10pt, conference]{IEEEtran}
\IEEEoverridecommandlockouts
\usepackage{cite}
\usepackage{amsmath,amssymb,amsfonts}
\usepackage{algorithmic}
\usepackage{graphicx}
\usepackage{textcomp}
\usepackage{xcolor}
\def\BibTeX{{\rm B\kern-.05em{\sc i\kern-.025em b}\kern-.08em
    T\kern-.1667em\lower.7ex\hbox{E}\kern-.125emX}}
\begin{document}

\title{HyperPalm: DNN-based hand gesture recognition interface for intelligent communication with quadruped robot in 3D space}

\makeatletter
\newcommand{\linebreakand}{%
  \end{@IEEEauthorhalign}
  \hfill\mbox{}\par
  \mbox{}\hfill\begin{@IEEEauthorhalign}
}
\makeatother

\author{
    \IEEEauthorblockN{Elena Nazarova}
    \IEEEauthorblockA{\textit{Intelligent Space Robotics Laboratory} \\
    \textit{Skoltech}\\
    Moscow, Russian Federation \\
    elena.nazarova@skoltech.ru}
\and
    \IEEEauthorblockN{Ildar Babataev}
    \IEEEauthorblockA{\textit{Intelligent Space Robotics Laboratory} \\
    \textit{Skoltech}\\
    Moscow, Russian Federation \\
    ildar.babataev@skoltech.ru}
\and
    \IEEEauthorblockN{Nipun Weerakkodi}
    \IEEEauthorblockA{\textit{Intelligent Space Robotics Laboratory} \\
    \textit{Skoltech}\\
    Moscow, Russian Federation \\
    nipun.weerakkodi@skoltech.ru}
\linebreakand
    \IEEEauthorblockN{Aleksey Fedoseev}
    \IEEEauthorblockA{\textit{Intelligent Space Robotics Laboratory} \\
    \textit{Skoltech}\\
    Moscow, Russian Federation \\
    aleksey.fedoseev@skoltech.ru}
\and
    \IEEEauthorblockN{Dzmitry Tsetserukou}
    \IEEEauthorblockA{\textit{Intelligent Space Robotics Laboratory} \\
    \textit{Skoltech}\\
    Moscow, Russian Federation \\
    d.tsetserukou@skoltech.ru}
}
\maketitle

\begin{abstract}
Nowadays, autonomous mobile robots support people in many areas where human presence either redundant or too dangerous. They have successfully proven themselves in expeditions, gas industry, mines, warehouses, etc. However, even legged robots may stuck in rough terrain conditions requiring human cognitive abilities to navigate the system. While gamepads and keyboards are convenient for wheeled robot control, the quadruped robot in 3D space can move along all linear coordinates and Euler angles, requiring at least 12 buttons for independent control of their DoF. Therefore, more convenient interfaces of control are required.

In this paper we present HyperPalm: a novel gesture interface for intuitive human-robot interaction with quadruped robots. Without additional devices, the operator has full position and orientation control of the quadruped robot in 3D space through hand gesture recognition with only 5 gestures and 6 DoF hand motion.

The experimental results revealed to classify 5 static gestures with high accuracy (96.5\%), accurately predict the position of the 6D position of the hand in three-dimensional space. The absolute linear deviation Root mean square deviation (RMSD) of the proposed approach is 11.7 mm, which is almost 50\% lower than for the second tested approach, the absolute angular deviation RMSD of the proposed approach is 2.6 degrees, which is almost 27\% lower than for the second tested approach. Moreover, the user study was conducted to explore user's subjective experience from human-robot interaction through the proposed gesture interface. The participants evaluated their interaction with HyperPalm as intuitive (2.0), not causing frustration (2.63), and requiring low physical demand (2.0).
\end{abstract}

\begin{figure}[!h]
\centering
 \includegraphics[width=0.9\linewidth]{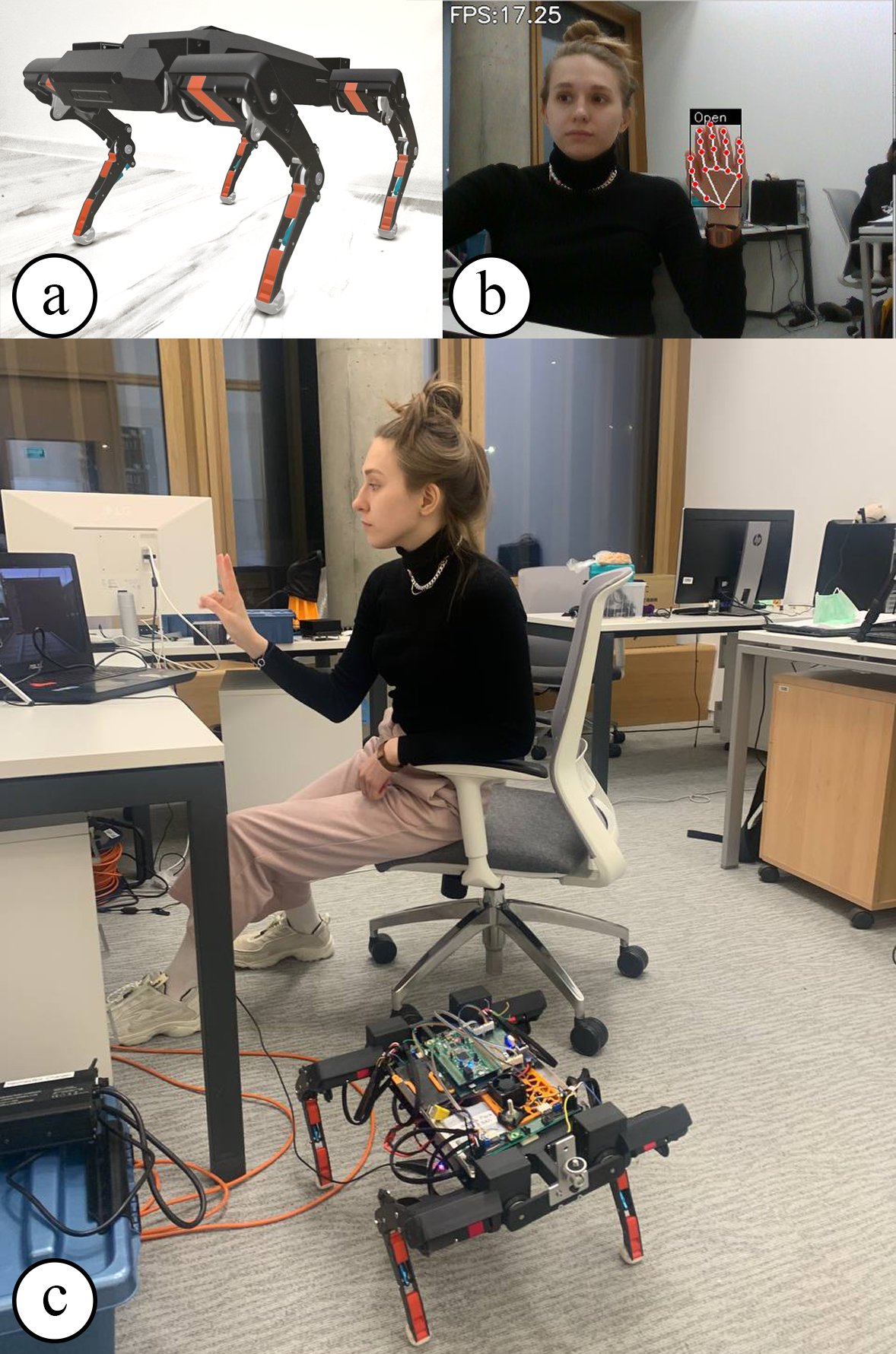}
 \caption{ (a) CAD design of the HyperPalm Robot. (b) DNN-based gesture recognition of the operator's hand. (c) HyperPalm Robot position and orientation control.}
 \label{fig:main}
\vspace{-1.5em}
\end{figure}

\section{Introduction}
Legged robots are increasingly being applied in the indoor and outdoor scenarios of exploration, delivery, and interaction with humans \cite{Biswal_2020}. While achieving lower velocities on planar surfaces compared with wheeled mobile robots, they have the potential to locomote on irregular terrains and navigate in cluttered environments. For example, Hiller et al. \cite{Hiller_2004} presented a quadruped robot design and control for an unstructured environment. Barasuol et al. \cite{Barasuol_2013} introduced a reactive controller for quadrupedal locomotion on challenging terrain. Self-organized locomotion with neural control was explored by Sun et al. \cite{Sun_2018}, where researchers demonstrated a successful simulation of a legged robot on flat terrain and in the presence of low-height obstacles.

Static gait for quadruped robots walking on uneven terrains, such as stairs, was investigated by Li et al. \cite{Li_2019} and Ye et al. \cite{Ye_2021}. The researchers achieved high results in the simulation of passability for the developed quadruped robots. The stair-climbing robot dog was also proposed by Campos et al. \cite{Campos_2019}, where the robot utilized CNN for both object detection and hand gesture recognition as part of its HRI strategy. Saputra et al. \cite{Saputra_2021} proposed an adaptive quadruped robot inspired by domestic felines, that was supporting advanced terrain climbing. Thus, the versatility of legged robots and quadruped robots, in particular, could potentially allow them to support several crucial tasks, in addition to the exploration and industrial e.g., navigation of people with sight disadvantages, explored with the guide robot dogs developed by Chuang et al. \cite{Chuang_2018} and Xiao et al. \cite{Xiao_2021}, or promote social distancing in crowded urban environments in the COVID-19 pandemic situation, as suggested by Chen et al. \cite{Chen_2021}. 

However, the live motion control of a four-legged robot is still a significant challenge in the robotic control field. Hence, many successful attempts of their automotive motion are by this day only demonstrated in simulations, while navigation of robots in the dynamic environment is supported by human operators. 


\section{Related Works}

With the emergence of the CNN-based and DNN-based approaches, the visual recognition of hand gestures has been significantly improved, finding their application in various scenarios of remote control. For example, a multi-sensor system for recognition of the driver’s hand motion was suggested by Molchanov et al. \cite{Molchanov_2015}, where the RGBD camera, and near-field radar data were combed for higher stability. Stancic et al. \cite{STANCIC_2017} proposed an inertial-based wearable system, that can apply hand-gesture dynamics for robotic control over high distances. Wearable gesture interfaces based on flex sensors were suggested by Afzal et al. \cite{Afzal_2017} for the control of robotic end-effector with four fingers and by Fedoseev et al. \cite{Fedoseev_2021} for the 6 DoF robotic arm control in drone catching task. 


Moreover, many works are currently devoted to the study of the interaction between a human and a swarm of unmanned aerial vehicles (UAV) and mobile robots through an interface based on gesture recognition. For example, a gesture interface developed Tsykunov et al. \cite{Tsykunov_2019} based on impedance swarm control and haptic feedback for HRI. Chen et al. \cite{Chen_2020} proposed the HRI multichannel robotic system in augmented reality (AR). A control approach with human hand arm and motions, which are recorded by a wearable armband, controlling a swarm’s shape and formation was suggested by  Suresh et al. \cite{Suresh_2019}. Alonso-Mora et al. \cite{Alonso-Mora_2015} and Kim et al. \cite{Kim_2020} suggested real-time input interfaces with swarm formation control. However, their approach was developed only for mobile robot operation in 2D space. 


\section{General system overview}

The developed HyperPalm system includes two Intel RealSense d435 RGB-D cameras, NVIDIA Jetson Nano on-board personal computer (PC) and operator PC located remotely, and HyperPalm robot itself with low-level communications (Fig. \ref{fig:Overview}).

\begin{figure}[htbp]
\centerline{\includegraphics[width=1.0\linewidth]{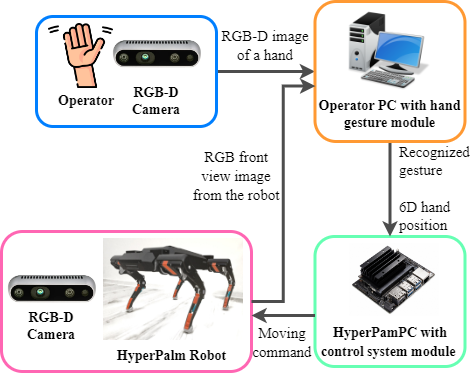}}
\caption{HyperPalm System Overview.}
\label{fig:Overview}
\end{figure} 

The first camera is located on the robot and captures the RGB image of its front view. This image is used for the operator to have understanding of the robot localization in the environment. The system is aimed at solving the problem of teleoperation when the robot performs the task remotely, and it cannot achieve the help from operator. Therefore, the operator needs visual feedback from the robot. The second camera captures the RGB-D image of the operator's hand. Both cameras send their output to the operator's PC. The robot's camera uses a wireless connection via open-source robotics middleware suite Robot Operating System (ROS). The operator camera uses a wired connection via a Universal Serial Bus (USB) cable. The operator's PC processes the input data through the hand gesture module, which will be described in the Hand Gesture Module Overview section. This module sends control data to the onboard computer NVIDIA Jetson Nano via ROS. Next, the onboard computer processes the input data from gesture module in the control system module and sends movement commands to the robot itself.

\subsection{HyperPalm Robot Design}

HyperPalm was designed and assembled using light 3D-printed and carbon fiber parts. Moreover, instead of the high-cost brushless direct current (DC) motors it utilizes DC servo motors, which allow HyperPalm to accomplish a high-precision movement. The leg motion is supported by the dampers, which can be seen in Fig. \ref{fig:Leg} to achieve a smooth transition of its feet on slippery and uneven surface.

\begin{figure}[h]
 \centerline{\includegraphics[width=0.5\textwidth]{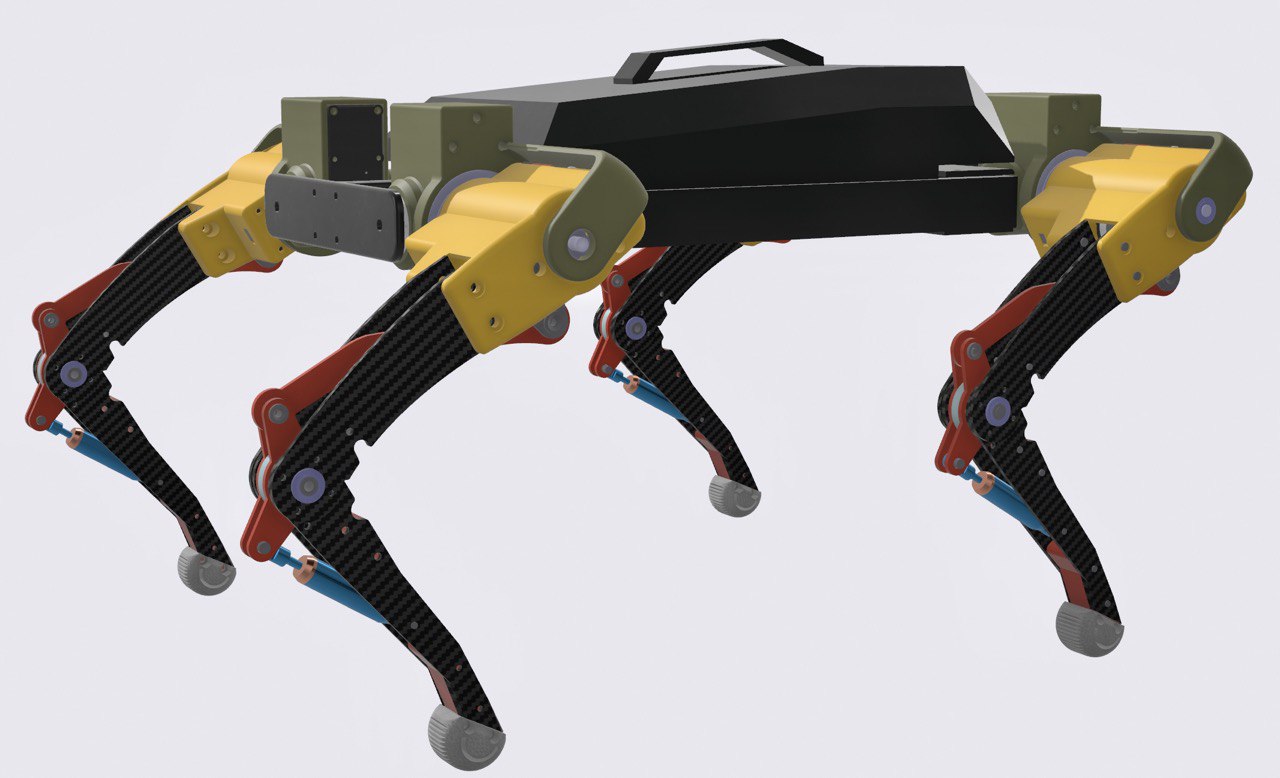}}
 \caption{HyperPalm Robot CAD design.}
 \label{fig:Leg}
\end{figure}

HyperPalm is a 12-DoF legged robot with (WxHxD: 300x175x240mm) external dimension. Each leg of the robot consists of 3 joints for hip, upper and lower legs. This helps to achieve a wide range of capabilities for robot movements. The robot itself has 5 kg weight and 2 kg pay-load capability. The robot is powered by an 8.4V and 8.8Ah Li-Ion battery pack, which allows the robot to run for about 30 minutes on a charge.

\section{Hand Gesture Module Overview}
\subsection{Gesture Recognition}
To switch the operating modes of the robot, proposed to use 5 static gestures: "One", "Two", "Three", "Open",  and "Close". This is necessary so that when the system is turned on and the operator's hands or others fall into the camera field of view, the robot does not react to the given changes in the position and orientation of the hand without specific run command.


We use DNN-based gesture recognition to achieve high precision in human gesture classification (Fig. \ref{fig:dnn}). DNN is a fully connected network with four layers: input layer, two hidden layers, output layer. Morower,  between layers the network has Rectified Linear Unit (ReLU) non-linear function, and batch normalization. The input layer takes 62 neurons, the first hidden layer has 256 neurons, the second hidden layer has 128, the last third output layer has 5 neurons with probabilities for each gesture class.

\begin{figure}[!h]
 \includegraphics[width=0.5\textwidth]{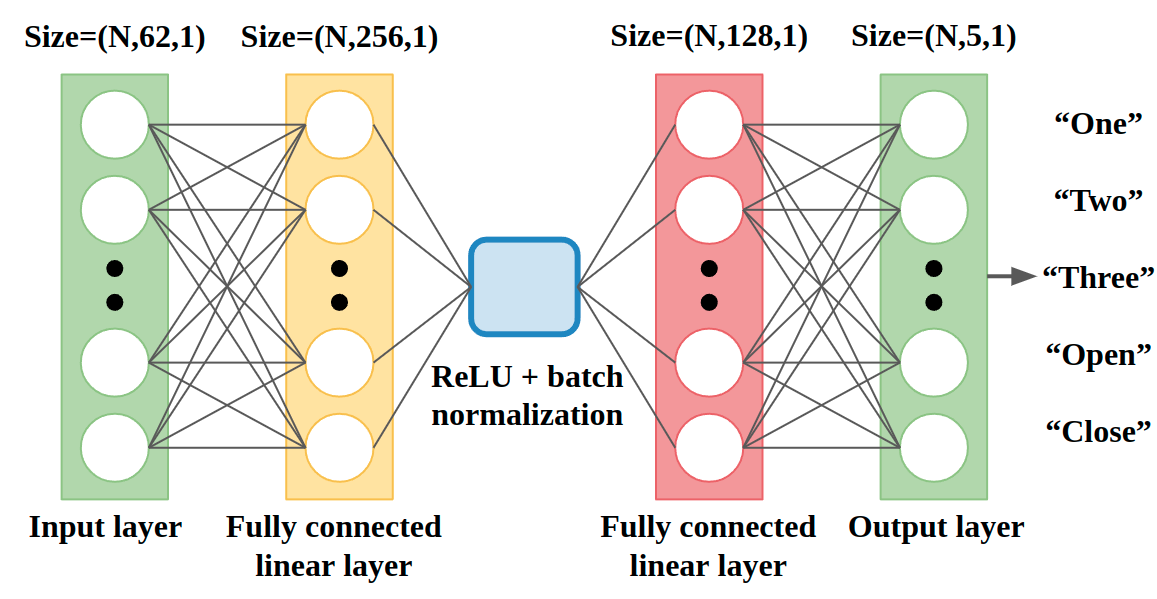}
 \caption{DNN model for gestures classification.}
 \label{fig:dnn}
\end{figure}

A gesture dataset for the model training consists of five gestures of 2500 arrays per each gesture. In total 12500 arrays with normalized $X$, $Y$ coordinates of 21 hand landmarks and 20 normalized vectors length between 21 hand landmarks. We divided dataset on train and test sets with a ratio of 75\% (9375 arrays) and 25\% (3125 arrays), respectively. It resulted in accuracy of 96.5\% when performing validation on a test set, which is shown in Fig. \ref{fig:Training accuracy}. .


\begin{figure}[!h]
\centerline{\includegraphics[width=0.5\textwidth]{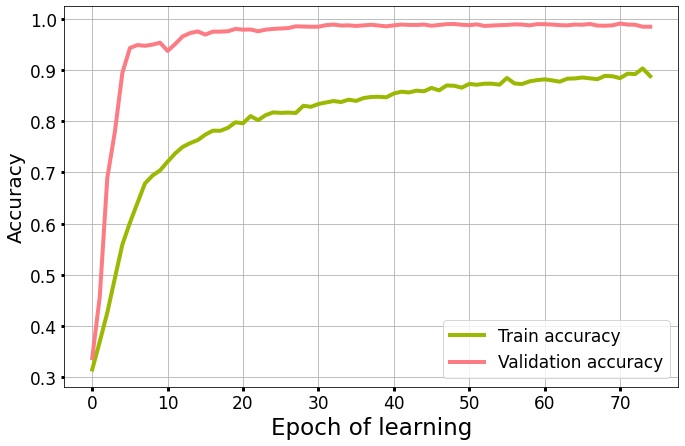}}
 \caption{DNN accuracy function on train and validation datasets.}
 \label{fig:Training accuracy}
\end{figure}

The model is invariant of the position and orientation of the human hand. If the probability of a gesture is less than 85\%, the algorithm takes this predicted gesture as the "None".

\subsection{Hand 6D Pose Estimation}
Determining the position of three linear coordinates of the hand, implemented by calculating relative translation of a hand in 3D space. Foremost, the algorithm defines a point in a space relative to which the deviation is calculated.  The 21 $x$ and $y$ coordinates of landmarks clearly receives from MediaPipe framework in pixels, which are converted to meters. The information obtains about the $z$ coordinate for each landmark of the hand from a pre-calibrated depth image in meters. Further, the algorithm calculates the 3D center of mass and analyzes its linear movement in space. It is necessary to obtain the entire hand and not 21 landmarks isolated.

The algorithm for calculating hand Euler angles consists of three parts.
Firstly, we extract $x$, $y$, $z$ coordinates of 21 hand landmarks as for determining the position of three linear coordinates. In addition, we implemented an algorithm to generate a point cloud of the hand via k-nearest neighbor approach. This is justified by the fact that the accuracy of expanded point cloud approximation to a plain is higher. On average, we increase the number of points in point cloud from 21 to 2000.

The next step is an approximation of the point cloud to a plane by least squares approach. The equation for a plane is shown in: 

\begin{equation}
ax + by + c = z\label{eq1}
\end{equation}
where $a$, $b$, $c$ are the coefficients of the scalar equation of the plane. 
Thus, we set up matrices with $x$, $y$, $z$ coordinates for each point in point cloud as follows:

\begin{equation}
\begin{bmatrix}
x_{0} & y_{0} & 1
\\x_{1} & y_{1} & 1
\\
 \vdots &  \vdots  &  \vdots
\\x_{n} & y_{n} & 1
\end{bmatrix}  \begin{bmatrix}a \\ b \\c  \end{bmatrix} =  \begin{bmatrix}z_{0}  \\z_{1} \\ \vdots \\z_{n} \end{bmatrix} \label{eq2}
\end{equation}
where $n$ is the number of points in pointcloud and is equal 2000 in our system. Since there are always more than three points in a point cloud, the system is over-determined. Therefore, we used the left pseudo inverse matrix as given in: 

\begin{equation}
\begin{bmatrix}a \\ b \\c  \end{bmatrix} =  (A^{T}A)^{-1}A^{T}B \label{eq3}
\end{equation}
where $A$ and $B$ are the matrices described by:

\begin{equation}
A = \begin{bmatrix}
x_{0} & y_{0} & 1
\\x_{1} & y_{1} & 1
\\
 \vdots &  \vdots  &  \vdots
\\x_{n} & y_{n} & 1
\end{bmatrix} \label{eq4} ,
\end{equation}

\begin{equation}
B = \begin{bmatrix}z_{0}  \\z_{1} \\ \vdots \\z_{n} \end{bmatrix} \label{eq5}
\end{equation}

Then we calculated the Euler angles of the plane in $x$, $y$, $z$ axes. The angle between two planes is equal to the angle determined by the normal vectors of the planes, as defined in:

\begin{equation}
Ang = Cos^{-1}(\frac{(a_{1}a_{2} + b_{1}b_{2} + c_{1}c_{2})}{(\sqrt{a_{1}^2+b_{1}^2+c_{1}^2)} (\sqrt{a_{2}^2+b_{2}^2+c_{2}^2)}})\label{eq6}
\end{equation}
where $a_{1}$, $b_{1}$, $c_{1}$, and $a_{2}$, $b_{2}$, $c_{2}$ are the direction ratios of normal to the first and second planes respectively.

\subsection{Control system to navigate HyperPalm}
In the proposed system, the operator can control the linear coordinates and Euler angles separately or together by switching the control mode. The operator shows a gesture called "One" to activate linear control mode. Next, the operator should show the "Open" gesture to initiate the calculation of the relative hand movements in 3D space. Finally, the operator demonstrates the "Close" gesture to disable 3D hand control immediately. The Euler angles control of the robot is activated by performing the "Two" gesture, combined linear and angular by "Three" gesture. The further control procedure is the same as for the linear coordinates control, Fig. \ref{fig:GestureInterface}.

\begin{figure}[htbp]
\centerline{\includegraphics[width=0.5\textwidth]{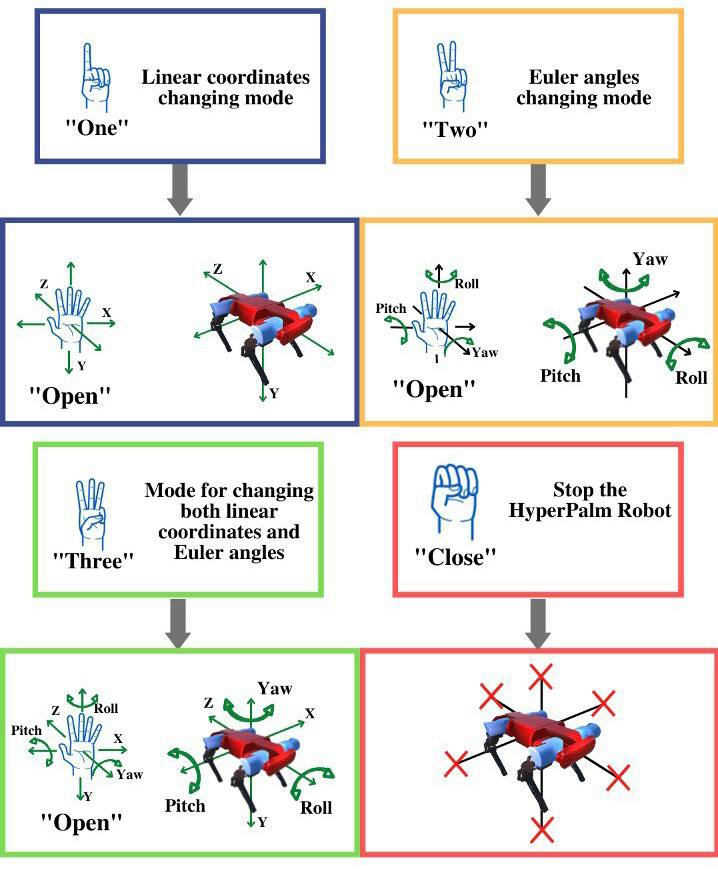}}
\caption{Control system architecture. Users control either position or orientation of the quadruped robot through the HyperPalm interface. The combinations of two gestures are used to swithch between control mods.}
\label{fig:GestureInterface}
\end{figure}

\section{Experiment: Human hand 6D pose estimation approach selection}
To choose a more accurate approach for determining the six-dimensional position of the hand in space, two approaches were tested and validated. The first approach is based on convolutional neural network (CNN), the second is the algorithm for calculating the 3D center of mass and approximating the hand point cloud to the plane, which was presented in the previous paragraph. 

The standard Root-Mean-Square Deviation (RMSD). metric was used to validate these approaches. It was chosen because RMSD can be interpreted as an absolute error. The RMSD represents the square root of the second sample moment of the differences between predicted values and observed values or the quadratic mean of these differences:

\begin{equation}
{RMSD} ={\sqrt {\frac {\sum _{t=1}^{T}({\hat {y}}_{t}-y_{t})^{2}}{T}}}
\label{eq7}
\end{equation}
where $T$ is a number of non-missing data points, ${\hat {y}}_{t}$ is estimated time series, and $y_{t}$ is actual observations time series.

\subsection{Yolo-based 6D hand pose estimation approach}
First approach based on work Tekin et al. \cite{tekin18} for simultaneously detecting an object in an RGB image and predicting its 6D pose without requiring multiple stages or having to examine multiple hypotheses. The key component of this method is CNN architecture inspired by works Redmon et al. \cite{inproceedings} and \cite{8100173} that directly predicts the 2D image locations of the projected vertices of the object’s 3D bounding box. The object’s 6D pose is then estimated using a Perspective-n-Point (PnP) algorithm.

To evaluate the hand posture, we collected our own dataset based on LineMOD benchmark for 6D object pose estimation, which consists of 2500 sequence capture RGB-D images, a processed binary masks for each image and a 3D model of a hand. We used an open-source project to create masks, bounding box labels, and 3D reconstructed object mesh for object sequences captured with an RGB-D camera. Moreover, we implemented a raw 3D model acquisition through aruco markers and ICP registration pipeline and processed it manually. In addition, the dataset was increased by adding the background augmentation using, 17000 random images from the Internet. 


\subsection{Proposed 6D hand pose estimation approach}
Second approach it is our method to estimate 6D hand pose based on the MediaPipe, depth maps and building point cloud for further processing.

Since for this approach the data were not pre-marked in advance, there is no ground for calculating RMSD. We invited 6 participants to determine the approximate ground truth. Each participant disposed in front of the UR3 from Universal Robots robot with a statically directed hand towards the camera, which is located on the robot (Fig. \ref{fig:ildar}). The robot moved along a pre-programmed trajectory along linear and angular axes, taken independently of each other. We discretely saved the linear and angular position of the robot and hand in millimeters (mm) and degrees. The repeatability of the UR3 is 0.1 mm. This is small enough to allow the robot's trajectory to be used as an approximate ground truth.

\begin{figure}[htbp]
\centering
\includegraphics[width=0.8\linewidth]{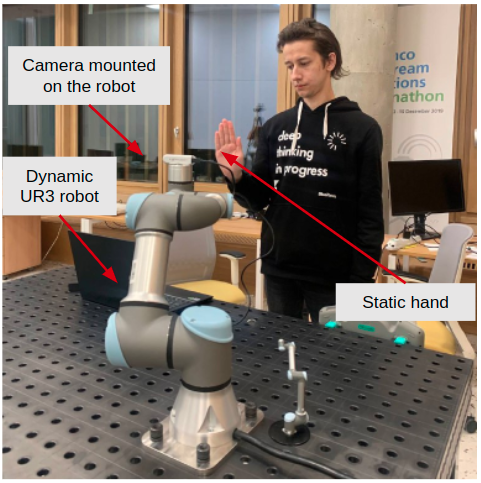}
\caption{The participant during the approximate ground truth estimation.}
\label{fig:ildar}
\end{figure} 

Thus, we obtained the trajectory of the robot, which, in the calculation for RMSD, is considered as ground truth and the trajectory of the hand, which was calculated by the algorithm for determining the 6D position of the hand, Fig. \ref{fig:lin}.

\begin{figure}[htbp]
\centering
\includegraphics[width=1\linewidth]{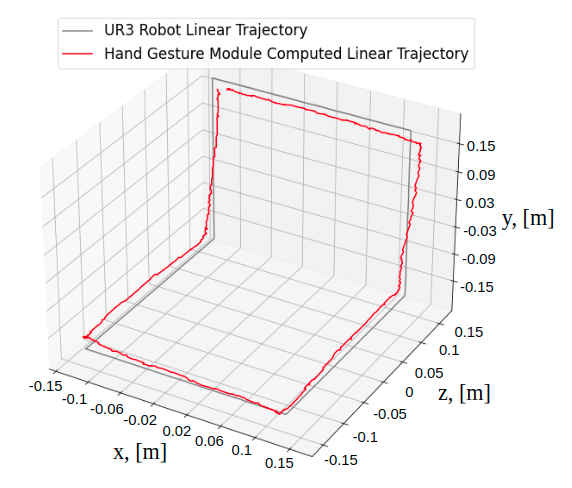}
\caption{The Hand gesture module computed a linear trajectory, and the UR3 Robot linear trajectory.}
\label{fig:lin}
\end{figure}

\subsection{Experimental Results}
We used the standard RMSD to compare 6D pose estimation error of two approaches in mm for linear and degrees for angular coordinates. 

The linear RMSD result for Yolo-based approach for all validation images in the best model the deviation is 23.6 mm. For our proposed approach for all participants who took part in the experiment the deviation is 11.7 mm, which is almost 50\% less than for the Yolo-based approach.

The angular RMSD result for Yolo-based approach for all validation images in the best model the deviation is 3.6 degrees. For our proposed approach for all participants who took part in the experiment the deviation is 2.6 degrees, which is almost 27\% less than for the Yolo-based approach.

According to RMSD result, our proposed approach has better performance to 6D hand pose estimation than Yolo-based approach. This is due to the fact that networks such as the Yolo-based depend on the shape changes of the object over time. We trained the algorithm for the human palm without specifically shape changing, but the human palm is exposed to external factors and its shape changes with time.  

\section{User experience evaluation}
The  goal of  this  paper is  to  evaluate the intuitive interaction between human and quadruped robot with the gestures recognition interface based on DNN. We provided an experiment to evaluate user experience with gesture-based control of HyperPalm system.

\subsubsection*{Participants}
We invited 8 participants (3 females) aged 19 to 27 years (mean = 23.0) to test HyperPalm system. Three of them have never interacted with gesture interfaces before, others were familiar with CV-based gesture interfaces. We covered student from different professional tracks.

\subsubsection*{Procedure} 
Participants were asked to execute several commands for gesture control. Before testing, they had a short briefing about test procedure and commands description. After the experiment, each participant completed a questionnaire based on The NASA Task Load Index (NASA-TLX) and three specific extra questions which give information such as age, gender of the participant, and how intuitive it was to control the robot with DNN-based gestures. The participants performed a training session, where each participant familiarized themselves with live interaction between themselves and HyperPalm with Gesture-based Interface. They were required to test several random commands and after recover initial position of robot.

An extra “intuitiveness" parameter was introduced in the survey, being an essential criterion in the teleoperation tasks to provide adjustable control over the swarm behavior in real-time.
Therefore, the participants provided feedback on seven questions.







 
\subsection*{Experimental Results}


The results of the NASA-TLX based survey are shown in Fig.~\ref{fig:likert}. 

\begin{figure}[htp]
\centering
\includegraphics[width=1\linewidth]{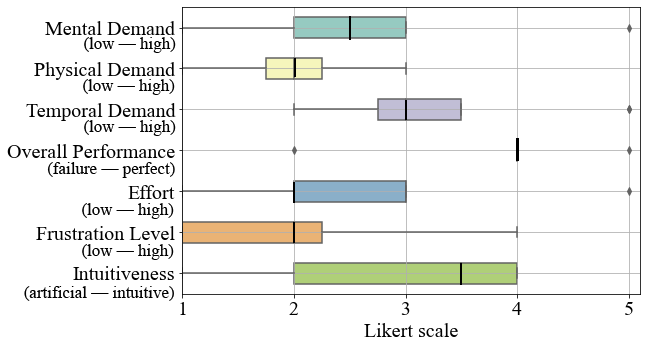}
\caption{Subjective feedback on the 5-point NASA-TLX based Likert scale.} \label{fig:likert}
\end{figure}

We conducted a chi-square analysis based on the frequency of answers in each category. The results showed that the parameters are all independent (min $p = 0.12 > 0.05$). 

In summary, six of the participants found the experiment with HyperPalm system exciting, whereas five did not feel any discomfort during control robot. The results revealed that participants were fully engaged in the gesture-based control and seven of them were not tired of robot operation procedure. All participants did not feel any additional physical effort during the gesture control performance (mean 2.0 out of 5.0).

\section{Conclusions and Future Work}
This paper introduced a novel DNN-based hand gesture recognition interface for intelligent communication with quadruped robot in 3D space. Where gesture recognition interface  allows the operator to guide the quadruped robot using 5 gestures and changing the position of the palm in space.

The HyperPalm interface provides immersive and intuitive control to the user. Using HyperPalm the participants’ feedback show that the interaction was mostly intuitive with a low frustration level (2.63 out of 5.0) and low physical demand (2.0 out of 5.0). The proposed human hand 6D pose estimation algorithm achieved the best results compare with Yolo-based 6D hand pose estimation approach. The linear RMSD is almost 50\% less, the angular RMSD is almost 27\% less.

In the future, we will provide the additional experiment with a scenario close to the real industrial environment. In this experiment, participants will be asked to navigate robot along a particular trajectory with several existing control interfaces and the developed the HyperPalm interface. Gesture recognition in various lighting levels and visual noise will be also experimentally evaluated.

\bibliographystyle{IEEEtran}
\bibliography{bibliography}
\end{document}